\documentclass[sigconf,nonacm]{acmart}
\usepackage{graphicx} % Required for inserting images
\usepackage{enumitem}
\usepackage{natbib}

\title[Introduction to Human-Robot Interaction: A Multi-Perspective Introductory Course]{Introduction to Human-Robot Interaction:\\ A Multi-Perspective Introductory Course}
\author{Tom Williams}
\affiliation{%
  \institution{MIRRORLab\\ Department of Computer Science\\ Colorado School of Mines}
  \city{Golden, CO}
  \country{USA}
}

\begin{abstract}
    In this paper I describe the design of an introductory course in Human-Robot Interaction. This project-driven course is designed to introduce undergraduate and graduate engineering students, especially those enrolled in Computer Science, Mechanical Engineering, and Robotics degree programs, to key theories and methods used in the field of Human-Robot Interaction that they would otherwise be unlikely to see in those degree programs. To achieve this aim, the course takes students all the way from stakeholder analysis to empirical evaluation, covering and integrating key Qualitative, Design, Computational, and Quantitative methods along the way. I detail the goals, audience, and format of the course, and provide a detailed walkthrough of the course syllabus.
\end{abstract}


\begin{document}

\maketitle

\section{Introduction and Course Goals}
In this paper I describe the design of an introductory course in Human-Robot Interaction. This project-driven course is designed to introduce undergraduate and graduate engineering students, especially those enrolled in Computer Science, Mechanical Engineering, and Robotics degree programs, to key theories and methods used in the field of Human-Robot Interaction that they would otherwise be unlikely to see in those degree programs. To achieve this aim, the course takes students all the way from stakeholder analysis to empirical evaluation, covering and integrating key Qualitative, Design, Computational, and Quantitative methods along the way.

The key goal of this course is three-fold. First, the course aims to introduce students to the notion of social robotics, and the idea of using interactive robots to help meet the needs of real world human communities. Second, the course aims to introduce students to the field of Human-Robot Interaction, and to showcase both the types of work that researchers are doing in the field to align with this vision of social robotics, as well as the research methods that HRI researchers used to achieve those goals. Finally, the course aims to operate as an HCI research methods course, where students can learn key tools, including qualitative research methodologies, design research methodologies, experimental design, and statistical analysis, which they could easily transfer to other engineering projects, regardless of whether they choose to pursue future work in HRI, or even in Computer Science at all.

\section{Prerequisites and Target Audience}

These course goals are critically conditioned on the expected background of the enrolled students. The course is offered at a small engineering-only university with a strong focus on Robotics related fields (50\% of all undergraduate students are enrolled in Mechanical Engineering or Computer Science degree programs, and degree programs offered in Robotics at both the undergraduate and graduate level), but with no degree programs offered in social sciences or humanities (e.g., Psychology) and few, if any, elective courses available in those fields. The university size and focus means that the course is offered at a mixed undergraduate/graduate level, and is primarily offered to students from Computer Science, Mechanical Engineering, and Electrical Engineering.
Based on this university structure and student makeup, students enrolled in the course have prerequisite programming and mathematical knowledge, have likely taken related technical courses such as Intro to Robotics, Robot Perception, SLAM, Robot Planning, Computer Vision, Robot Ethics, Mechatronics, Advanced Robotic Control, or Robot Mechanics, and have likely taken key design courses required of all undergraduate students, but likely have little exposure to or appreciation for relevant theories, methods, or practices from psychology, philosophy, communication, and so forth. 
These assets and deficits critically shape the content covered in the course.

One surprising way these assets and deficits shape course design is in terms of the course's coverage of robot ethics topics (or lack thereof). 
Most students enrolled in the course also take Robot Ethics either before or after taking this course. This is especially true for graduate students enrolled in a Robotics MS or PhD program, who are required to take one or both courses. As such, despite the societal and ethical impacts of interactive robots being absolutely critical to the course content, these topics are not explicitly covered since it is assumed that most students will receive deep coverage of those topics in the standalone Robot Ethics course. That being said, key ethical frameworks such as Engineering for Social Justice (E4SJ)e~\cite{leydens2017engineering}, Robots for Social Justice~\cite{zhu2024hri}, and Feminist Human-Robot Interaction~\cite{winkle2023hri} are baked into the course and reflected in course activities such as stakeholder analysis, needfinding, and E4SJ-grounded reflection exercises. Moreover, discussion of social and ethical implications frequently arise in guest lectures invited in the second half of the semester.

\section{Course Format Overview}
To achieve the course goals, the course is structured as a 48-student, project-based, sixteen-week course with an even balance between lectures and lab assignments. Several lab assignments and lectures are derived from reference courses, especially Dr. Ana Paiva's Masters-level \textit{Social Robots and Human Robot Interaction} course, and Bilge Mutlu's graduate-level \textit{Human-Computer Interaction} course, albeit adapted for this course's mixed undergraduate/graduate population.

All 48 students attend class Mondays and Wednesdays for 50 minutes each day. Students also attend one of two 24-student, two-hour long, lab section on Fridays. Course exercises use nine Softbank Naos purchased for the course over several years through university-internal course equipment grants. At any given time, course staff ensure that eight of the nine Naos are charged and operational. As such, each lab section can be broken into eight three-student project groups, each of which has a Nao made available to them during lab sections.

Over the course of the semester, students submit all lab reports and other project deliverables by uploading them to Open Science Framework (OSF) repositories, thus teaching the students open science principles and tools as a secondary learning outcome.

\section{Detailed Syllabus}

In this section I will go through the course syllabus week by week, to explain how the different elements of the course fit together.

The first week begins with a set of activities intended to convey a high level sense of the course to students. Specifically, the first class immediately establishes the importance of grounding robotic engineering practice in genuine community needs, and the ability of qualitative methods to establish this grounding. Students are then immediately started on their first major assignment, in which they are tasked with interviewing someone of their choice who works in an industry relevant to social robotics, about the needs they face. In this first week, a panel of HRI industry professionals also attend class, to help students see this first assignment, which is likely out of their comfort zone, as an opportunity to learn a skill relevant to their future robotics careers. Finally, in this first week, students participate in their first Nao programming lab, to better understand the capabilities of the robot they will use in the class, and better see connections with the needs of the stakeholders they are interviewing.

The second week of the course steps back to explore the theoretical foundations of HRI, including key dimensions of interactions and interactivity, and of longer-term constructs such as trust and influence. Finally, students end the week by performing a multi-day grounded theory analysis of their interview transcripts, modeled after an assignment from Bilge Mutlu's HCI course at University of Wisconsin Madison. In this assignment, students begin by performing open coding together in class. Students then separately perform axial coating as homework, and then reconvene in groups to consolidate their axial codes and identify higher level trends in their collected interview transcripts. During this week, students are also tasked with performing brief literature reviews to supplement what they are hearing in their interviews with what others have heard and determined in other parts of the field. 

In the third week of the class, class is devoted to collaborative exercises in which students work towards a shared understanding of the goals they will pursue over the course of the semester. Students begin by taking insights from their interviews, and insights from their literature reviews, and creating sets of need-reason-source triads that correspond with key user needs, associated reasons why those needs should be prioritized in robot design (if possible. And the source of that reason in their interview or literature review. This helps to ensure that student projects are grounded in real user needs and are traceable back to those needs. Next, students use these need-reason-source triads to create vision statements for their class projects. Third, students engage in structured reflection exercises in which students are encouraged to reflect on ethical design principles from Engineering for Social Justice~\cite{leydens2017engineering}, and are given the opportunity to revise their vision statements accordingly. Finally, students identify key robot design goals aligning with their final vision statements, and perform a card sorting exercise to identify which design goals they will actually prioritize over the course of the semester. 

\begin{figure}
    \centering
    \includegraphics[width=\linewidth]{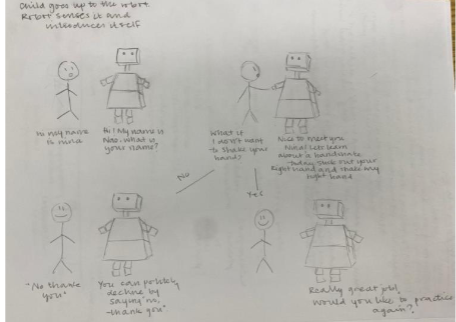}
    \caption{Storyboard from a project group in Fall 2022}
    \label{fig:storyboard}
\end{figure}

In the fourth week of the class, students are given the tools they need to pursue their design goals. We begin with a lecture on robot design, followed by a robot design lab in which students first storyboard out interactions (e.g., Fig.~\ref{fig:storyboard}), then learn principles of improvisational theater through Improv theater games played outside on the campus quad, then apply those principles to perform roleplay–based  embodied sketching exercises. These exercises help students identify possible interaction designs that will help them best achieve their design goals. Students then perform further literature review to identify other interaction design strategies from the literature that might also help them achieve their design goals. Finally, students end this fourth week with midterm presentations in which they present the results of their interviews, their design visions, design goals, and the design strategies they plan to use to pursue those goals and visions to meet the needs of their identified user populations.

In weeks five through eight, students learn the technical skills they need to execute their design strategies. 
In week five, we focus on spatial and non-verbal interaction. Students receive lectures on proxemics, motion, and gaze, and begin a lab assignment on robot perception, grounded in the official Nao tutorials.
In week six, we focus on interaction dynamics. Students receive lectures on group and team structures, collaboration, and turn-taking, and complete their perception lab assignment.
In week seven, we focus on verbal interaction. Students receive lectures on dialogue, reference, and grounding, and begin a lab assignment on robot dialogue, grounded in the official Nao tutorials.
Finally, in week eight, we focus on emotion and personality. Students receive lectures on emotion and character design, and finish their dialogue lab assignment.

In weeks nine through twelve, students are given the tools needed to run experiments to evaluate their designs. In week nine, students are taught about key dimensions of quantitative research methods, including measures, metrics, and experimental design. Students then begin a month-long experimental design lab, in which they formulate research questions grounded in their design strategies, formulate hypothesis that correspond with those research questions, and design a video-based HRI experiment to test those hypotheses. Students then use the design skills learned earlier in the semester to design the interactions that will appear in their experimental stimuli, and use the technical skills learned earlier in the semester to implement those designs on the Nao robot. Students go through CITI training, and create research ethics protocols and consent forms for their experiments. Finally, students implement their video-based experiments through Google Forms, and run their experiments, with each student in the class serving as a participant in each study designed by a project group other than their own. During weeks ten through twelve, students work on this lab in class together on Fridays and outside of classes homework, while on Mondays and Wednesdays, guest speakers from other universities provide guest lectures on a wide array of HRI topics. These guest lectures broaden the scope of HRI topics to which students are exposed, without adding additional assignments to the students workload. 

In weeks thirteen and fourteen, students are given the statistical tools needed to evaluate the data from the experiments they have run. In week thirteen students learn about both Frequentist and Bayesian statistics, and complete a two-week lab assignment, in which they first proceed through external tutorials on performing data analyses in JASP, and in which they are then invited to use the same analysis paradigms to analyze the results of their own experiments. In week fourteen, students listen to another guest lecture from an HRI researcher from another university while completing their statistics lab assignment, and then are released on Thanksgiving break.

Finally, we close the semester in week fifteen and sixteen. In these last two weeks, students hear a final invited talk from an external researcher, and perform a design futures exercise in which they speculate about other possible future uses for social robots beyond those they were able to explore in the class. Students attend a course wrap up lecture in which they learn about other classes they can take to supplement their knowledge, and learn about where other HRI research is being performed across the US, and especially in Colorado, in case they are interested in pursuing graduate work in HRI. Finally, students give group presentations covering the entire life cycle of their projects, from needfinding, to design, to implementation, to evaluation and results.

\section{Assignments and Assessment}

Over the course of the semester, students are assessed through several means.
All students in the course read key course readings (typically textbook chapters from \citet{bartneck2020human}'s \textit{Human-Robot Interaction: An Introduction}) and are assessed using brief, simple, low-stakes quizzes on the reading.
In addition, graduate students in the class are asked to read a series of HRI research papers relating to class topics, and are assessed using forum-style discussion-board posts in which they engage with the content of those papers.

Next, students are graded through project deliverables, including lab reports, midterm and final presentations, a final video demonstration of the autonomous interaction they designed for the Nao, and a final six-page HRI-style paper reflecting the type of paper they \textit{could} have submitted to HRI if they had actually obtained IRB approval and run the experiment with naive participants rather than pilot-testing the experiment on themselves. In previous years, I have had students actually obtain IRB approval, and paid for student groups to run twenty within-subject participants each on Prolific. However, in practice the overhead needed for students to learn to use psiTurk, and the overhead needed for the course staff to deploy and run these experiments on students' behalf, was deemed prohibitively burdensome and not sufficiently contributing to student learning.

Finally graduate students are asked to complete a four-page mini-survey on a topic of their choice within HRI, in which they collect, compare, and contrast at least 20-30 research papers on their chosen topic. At least half of these papers are required to have been published at T-HRI, HRI, RO-MAN, or ICSR.

\section{Case Study}
Each year, students in this course have explored a wide range of topics through their semester-length projects.
To provide a sense of these projects, I will discuss the outcomes of one undergraduate project team as a case study. Approval to provide this information was approved by our Human Subjects Research board, under informed consent from students.

This project team began by interviewing an elementary school teacher about her experiences in the classroom. Based on this interview, and inspired by the HRI 2018 paper ``Stop. I See a Conflict Happening''~\cite{shen2018stop} the students formulated the following design vision:

\begin{quote}
\textit{    ``A key goal in elementary classrooms identified was that students and teachers need to have individual interactions to assist with communication and manage conflicts. From the interview and axial codes, several issues that make it hard for teachers to achieve this goal were identified. These issues include that kids struggle with social interactions, get frustrated quickly, prefer iPads to social interaction, and teachers are often required to get involved in physical conflicts between students. In addition, meeting the needs of every student individually requires teachers and students to have one on one time. However, as seen in axial code, teachers are required to perform many varied and demanding classroom tasks and thus do not have the time or manpower to deal with and manage social interaction and conflict with students on an individual basis. We believe that social robots may be able to assist teachers in managing conflicts between students and help to teach positive social interaction. We believe that social robots address this problem because robots can interact one on one with students in scenarios where the teacher is busy, and can intervene in minor conflicts between students freeing the teacher to focus on other tasks. In addition, kids may be more willing to listen to advice and rules from a robot and be less likely to fight a robot.''}
\end{quote}

After formulating this power, students designed a conflict resolution interaction using storyboarding and embodied sketching, implemented it on the Nao robot using its Python API, and chose to study how different linguistic choices might shape the persuasive power of the robot as derived from its perceived authority. The students designed a 2 (Humorous/Serious) x 2 (Assertive/Passive) within-subjects experiment with a Latin Square design, and measured the effects of these conditions on perceived authority using the scale proposed by \citet{gudjonsson1989compliance}. Finally, the students analyzed their results under a Bayesian analysis framework, and calculated Bayes Inclusion factors for each of their considered factors and their interaction.

Overall, these activities demonstrated students' (undergraduate-level) mastery over qualitative, design, computational, and quantitative research methodologies.

\section{Conclusion}
Overall, \textit{Introduction to Human-Robot Interaction} serves to introduce students to the domain of Social Robotics, the field of Human-Robot Interaction, and the research methodologies of Human-Computer Interaction. Due to the cross-cutting nature of the course, the course does not delve deeply into (1) the fundamental theories of HRI, (2) the algorithmic methods of HRI, or (3) the social and ethical implications of HRI. Students taking this course (especially graduate students) would thus be best served by taking relevant courses in those areas before, concurrently with, or following the course.

\begin{acks}
    This work was funded in part by NSF CAREER Award IIS-2044865.
\end{acks}

\bibliographystyle{ACM-Reference-Format}
\bibliography{main}

\end{document}